\documentclass[pre,twocolumn,superscriptaddress,floatfix]{revtex4-1}  

\usepackage[paperwidth=210mm,paperheight=297mm,centering,hmargin=2cm,vmargin=2.5cm]{geometry}

\usepackage[pdftex]{graphicx}
\usepackage{amsmath,amsfonts,amssymb}
\usepackage{natbib}
\usepackage{MnSymbol}

\usepackage{hyperref}
\usepackage{xcolor} 
\usepackage{array}
\usepackage{pgfplots}
\usepackage{graphicx}
\usepackage[left,modulo,pagewise]{lineno}
\usepackage{dsfont}
\usepackage{amsmath}
\usepackage{graphicx}
\usepackage{rotating,wrapfig}
\usepackage{xspace}
\usepackage{xcolor} 

\usepackage{filecontents}

\definecolor{bluemoi}{rgb}{0.25,0.50 ,0.75} 

\hypersetup{
  bookmarksopen=true,
  pdftitle=" ",
  pdfauthor=" ", 
  pdfsubject=" ", 
  pdftoolbar=true, 
  pdfmenubar=true, 
  pdfhighlight=/O, 
  colorlinks=true, 
  pdfpagemode=UseNone, 
  pdfpagelayout=SinglePage, 
  pdffitwindow=true, 
  linkcolor=bluemoi, 
  citecolor=bluemoi, 
  urlcolor=bluemoi 
}

\newcommand{\Sr}{S_{row}}
\newcommand{\Sc}{S_{col}}
\newcommand{\Sb}{S_{ch}}
\newcommand{\tf}{TF\xspace}
\newcommand{\mt}{\textit{M3}\xspace}
\newcommand{\disc}[1]{\textcolor{black}{#1}}
\newcommand{\discf}[1]{\textcolor{black}{#1}}

\newcommand{\titre}{A framework for remote sensing images processing using deep learning techniques}

\makeatletter
\renewcommand{\figurename}{\sf \textbf{Figure}}
\renewcommand{\thefigure}{\arabic{figure}}
\renewcommand{\fnum@figure}{\sf\textbf{\figurename}~\textbf{\thefigure}}
\renewcommand{\tablename}{\sf\textbf{Table}}
\renewcommand{\thetable}{\arabic{table}}
\renewcommand{\fnum@table}{\sf\textbf{\tablename}~\textbf{\thetable}}
\makeatother

\begin{document}

\title{\titre}
\author{R\'emi~Cresson}
\thanks{R. Cresson is with the IRSTEA, UMR TETIS, BP 5095, Montpellier, France. remi-dot-cresson-at-irstea.fr}

\begin{abstract}
Deep learning techniques are becoming increasingly important to solve a number of image processing tasks. 
Among common algorithms, Convolutional Neural Networks and Recurrent Neural Networks based systems achieve state of the art results on satellite and aerial imagery in many applications.
While these approaches are subject to scientific interest, there is currently no operational and generic implementation available at user-level for the remote sensing community.
In this paper, we presents a framework enabling the use of deep learning techniques with remote sensing images and geospatial data.
Our solution takes roots in two extensively used open-source libraries, the remote sensing image processing library Orfeo ToolBox, and the high performance numerical computation library TensorFlow.
It can apply deep nets without restriction on images size and is computationally efficient, regardless hardware configuration.
\end{abstract}

\maketitle

\section{Introduction}
The volume of earth observation data never stopped to increase in recent years. 
In one hand, the use of very high spatial resolution satellites (VHRS) images is booming, and sensors are each generation sharper~\cite{BEN}.
There is also more and more sensors with close temporal acquisition, such as the Sentinel constellation.
In the other hand, community-based geographic information gathering systems are expanding each years in-situ geospatial data bases like OpenStreetMap~\cite{OSM}.
Deep learning (DL) is a growing trend in big data analysis, and had a breakthrough impact in the last few years on such diverse domains like Computer Vision, Speech Recognition, Image Processing, and Remote Sensing (RS). 
Convolutional Neural Networks (CNN) are designed to extract features in images, enabling image recognition, object detection, semantic segmentation. 
Recurrent Neural Networks (RNNs) are suited for sequential data analysis, such as speech recognition and action recognition tasks. 

In recent years, a number of studies have shown that RS benefits strongly from these new approaches, thanks to the availability of data and computing resources~\cite{ZHU}. 
Many typical RS problems have been successfully addressed with these architectures:
synthetic aperture radar interpretation with target recognition~\cite{CHENSAR}, classification from time series~\cite{MINH}, and parameter inversion~\cite{WANG16},
hyperspectral image analysis with classification~\cite{CHEN14}, anomaly detection~\cite{LI17},
VHRS images interpretation with scene classification~\cite{VOLP,MAGG}, object detection~\cite{CHENG16}, image retrieval~\cite{NAPO}, classification from time-series~\cite{IENC2}.
DL has addressed other issues in RS, like data fusion (see \cite{GOM15} for a review) e.g. multimodal classification~\cite{IENC}, pansharpening~\cite{SCA}, and 3D reconstruction.

DL allows researchers in RS to move beyond usual approaches and tackle a number of problems with solid results. 
However, there is still challenges remaining ahead in operational context. 
First, one crucial point is software implementation.
Entering into DL requires to familiarize with a framework. 
Several popular open-source frameworks exist, such as Caffe, Torch, Theano, and TensorFlow, but implementing new methods requires \disc{an extensive programming background and DL knowledge}, holding back DL democratization in RS community.
Conversely, many studies rely exclusively on pre-processed, densely annotated public datasets like UC Merced~\cite{MERCED} or Postdam~\cite{POSTDAM} because data extraction requires knowledge of geospatial standards and tools like Geographic Information Systems and RS image processing software.
Secondly, the need of tools that scale enough to the huge size of real-world RS images and datasets, raises as a major computing challenge.
While RS images processing software have internal mechanisms to deal with the large stacks of images, this problematic is out of DL frameworks scope.
Another point is that algorithms presented in the literature are often serving one unique purpose.
Therefore, coding is not generic and each new algorithm implementation require new effort and expertise.
Finally, recent studies have pointed that using the features computed with deep nets in machine learning algorithms like Support Vector Machines (SVM) or Random Forests, offers state of the art results~\cite{NOGU,MARM,HU15}. 
However, even if RS image processing software implement generally a machine learning framework, there is no existing one enabling the combination of DL algorithms with already implemented well-known machine learning algorithms.

Regarding these matters, we propose a generic framework to enable DL for RS images, which (i) is open-source and cross-platform, \disc{(ii) enables users without programming knowledge to use deeps nets on RS images, (iii) is integrated in an existing rich machine learning framework for RS images, (iv) is computationally efficient and allows the processing of very large RS images}.
Our approach is based on the RS image processing library, Orfeo Toolbox (OTB)~\cite{OTB} and rely on the high performance numerical computation library for training and inference on deep neural networks, TensorFlow (\tf)~\cite{TENSF}.
The existing machine learning framework of OTB includes tools for RS data selection, sampling, training, classification, regression, with many strategies and parameters fine tuning.
\disc{We extend this framework using \tf to enable the processing of RS images with deep nets.}
We first give in Section \ref{sec_method} a detailed description of our method. \disc{We then analyze the scalability of our approach} with state of the art deep nets and Spot-7 and Sentinel-2 images (Section \ref{sec_Expe}). We finally discuss the main advantages and limits of our approach in Section \ref{sec_disc}.

\section{Method}
\label{sec_method}

\subsection{Overview}
\label{sssec_Overview}

The Orfeo ToolBox (OTB) is a library for RS image processing, built on top of an application development framework widely used in medical image processing, the Insight Toolkit (ITK)~\cite{ITK}.
The machine learning framework of OTB is able to process large datasets at continental scale for land mapping~\cite{INGLADA} and benefits from High Performance Computing (HPC) architectures like clusters~\cite{CRES}.
TensorFlow (\tf) is a library for dataflow programming. 
It is a symbolic math library, and is intensively used for machine learning applications such as deep nets. 
It can also operate at large scale in HPC architectures like GPU.
We aim to extend the existing rich machine learning framework of OTB with \tf.
Implementation of state of the art deep nets should be enabled with the minimum effort, and the opportunity to apply them on RS images must be granted to non-developers, namely users.
The existing user-oriented machine learning framework of OTB must be preserved. We also sought the implementation of a component which can be used in a transparent way inside any OTB pipeline, to enable the combination of already implemented approaches with DL.
In the following sections, we provide description of the libraries processing frameworks (Sections \ref{ss_Libraries} and \ref{ss_Pipelining}). Then we introduce our low-level OTB compliant component in Section \ref{ss_ModelFilter} and our new user-oriented OTB applications in Section \ref{ss_NewApps}. Finally, we analyze performances of our approach with state of the art RS deep nets in Section \ref{sec_Expe}.

\subsection{Libraries paradigms}
\label{ss_Libraries}

This section describes the processing logic of OTB and \tf.

\subsubsection{OTB workflow}
\label{ssss_ITK-OTB}

A \textit{pipeline} refers to a directed graph of process objects,  that can be either sources (initiating the pipeline), filters (processing the data) or mappers (typically, write a result on disk).
Sources and filters can generate one or multiple data objects (e.g. image, vector). 
In the same way, filters and mappers consume one or multiple data objects.
The architecture of OTB inherits from ITK and hides the complexity of internal mechanisms for pipeline execution, which involve several steps. 
\disc{Detailed description of the pipeline mechanism can be found in~\cite{OSA} including figures and sequence diagram.}
The execution of a pipeline starts from a mapper, triggering its upstream filter(s).
When a filter is triggered, information about mandatory input data (i.e. information about output data of upstream process object(s)) is also requested upstream. 
In this way, it is propagated back through the pipeline, from mappers to sources via filters. 
Once this request reach sources, data objects information are generated (e.g. image size, pixel spacing) then propagated downstream to mappers. 
It should be noted that filters can potentially modify these information, according to the process they implement (e.g. changing image size) which is the case with image resampling for instance. 
Finally, they reach the mapper, initiating the data processing. 
Information regarding the size of the image that will be produced, is then used by the mapper to choose a splitting strategy. 
The default splitting scheme is based on the system memory specification. Other strategies can be chosen, e.g. striped or tiled regions with fixed dimensions. 
Then, the mapper requests its input image to upstream filter(s) sequentially, region after region. 
The data request and generation is handled through the pipeline in the same way as for the information: once the request reaches the sources, initiating the pipeline, the requested region is produced, then processed through filters, to finally end in the mapper. The pipeline execution continues with the next image region until the entire output is generated. This mechanism, named \textit{streaming}, enables the processing of very large images regardless the memory consumption of the process objects composing the pipeline.

\subsubsection{\tf workflow}
\label{ssss_ITK-OTB}

\tf uses symbolic programming, which distinguish definition of computations from their execution.
In \tf, tensors are abstraction objects of the operations and objects in the memory, simplifying manipulation regardless the computing environment (e.g. CPU or GPU).
A \tf model consists in operations arranged into a graph of nodes.
Each node can viewed as an operation taking zero or more tensors as inputs, and producing a tensor.
This data flow graph defines the operations (e.g. linear algebra operators), but the computations are performed within the so-called \textit{session}.
A high-level API enables the construction of \tf graphs, and the session runs the graph in delegating calculations to low level, highly-optimized routines.
Among tensors, we can distinguish concepts such as \textit{Placeholders}, \textit{Constants} and \textit{Variables}. A Placeholder is a symbol hosting input data, e.g. a set of images. As its name indicates, Constants are tensors with constant values, and Variables hold non persistent values, e.g. parameters to estimate during training. Variables must be explicitly initialized, and can be saved or restored during a session along with the graph.
A number of tools enable design of \tf models, like the \tf Python API or user-oriented software developed by the \tf community.

\subsection{Flowing the pipeline}
\label{ss_Pipelining}

In this section, we present prerequisites for the integration of a process object that runs \tf session for generic deep nets, in a OTB pipeline with RS images as data objects. 
In the pipeline workflow, the generation of images information and the requested regions computation by process objects, are crucial steps.
We denote the \textit{spacing} the physical size of a single RS image pixel.
The generation of output images information includes origin, spacing, size, and additional RS metadata e.g. projection reference.
Process objects must also propagate the requested regions to input images.
Regarding deep nets implementation, and particularly CNNs, this process must be carefully handled. 
CNNs usually involve several operations, mostly including a succession of convolutions, pooling, and non-linear functions. 
There is also a number of deriving operators e.g. transposed convolution. 
Most of these operators modify the size and spacing of the result.
For example, convolution can change the output image size, depending its kernel size and input padding. 
It can also change the spacing of the output if performed with strides, that is, the step of which is shifted the filter at each computation. 
Pooling operators are also a common kind of operator that modify the output size and spacing, depending of the stride and the size of the sliding window. 
We should note that a number of other operators change size and scale the spacing of the output.

Considering a process object implementing such operations, it must propagate requested regions of images to its inputs.
In the following, we introduce a generic description of size and spacing modifications that a deep net induces in processing RS images.
We name the \textit{scaling factor} of a particular output of the net, the ratio between the output spacing and a reference spacing (typically one input image feeding the deep net).
This parameter enables the description of any change of physical size of the pixel introduced by operators such as pooling or convolution involving non-unitary strides.
In addition, each input has its own \disc{\textit{receptive field}}, the input space that a particular output of the net is affected by. 
In the same way, each output has its own \textit{expression field}, the output space resulting from the \disc{receptive field}.
Images regions and spacing modifications induced by a \tf graph are thus defined with the \disc{receptive field} of inputs, and the scaling factor and expression field of outputs.

\subsection{Running \tf session in process object}
\label{ss_ModelFilter}

We introduce a new OTB process object which internally invokes the \tf engine and enforces the pipeline execution described in Section \ref{ss_Pipelining} \disc{and hence, }\discf{can seamlessly process large images}.
It takes one or multiple input images, and produces zero or multiple output images.
The input placeholders of the \tf model are fed with the input images of the process object.
After the session run, computed tensors are assigned to the process object outputs.
Placeholders corresponding to input images, as well as tensors corresponding to output images, are both specified as named in the \tf model.
The process object uses \disc{receptive field} of inputs, as well as expression field and scaling factor of outputs, to propagate requested input images regions and generate information of output images.
Finally, two processing modes are currently supported: 

\subsubsection{Patch-based}
Extract and process patches independently at regular intervals.
Patches sizes are equal to the \disc{receptive field} sizes of inputs. 
For each input, a tensor with a number of elements equal to the number of patches feds the \tf model.

\subsubsection{Fully-convolutional}
Unlike patch-based mode, it allows the processing of an entire requested region.
For each input, a tensor composed of one single element, corresponding to the input requested region, is fed to the \tf model.
This mode requires that \disc{receptive fields}, expression fields and scale factors are consistent with operators implemented in the \tf model, input images physical spacing and alignment. \discf{Blocking artifact is avoided by computing input images regions aligned to the expression field sizes of the model, and keeping only the subset of the output corresponding to the requested output region.}

\subsection{New OTB applications}
\label{ss_NewApps}

OTB applications generally implement pipelines composed of several process objects connected together.
We first provide a new application dedicated to RS images sampling suited for DL.
Then, using our new filter described in Section \ref{ss_ModelFilter}, we provide new applications for \tf model training and serving.
Finally, we introduce some new applications consisting in an assembly of multiple OTB applications.
All these new applications are integrated seamlessly in the existing machine learning framework of OTB.

\subsubsection{Sampling}
\label{sss_AppSampling}

The existing machine learning framework of OTB includes sample selection and extraction applications suited for geospatial data like vector layers and RS images. 
Multiple sampling strategies can be chosen, with fine control of parameters. 
However, DL on images usually involves reference data made of patches, but the existing application performs pixel-wise sample extraction.
We introduce a new application that performs patches extraction in input images, from samples positions resulting from the existing sample selection application of OTB.
As we want to stick to RS images files formats, samples are concatenated in rows to form one unique big image of patches. 
Typically, considering a number $n$ of sampled patches of size $[\Sr,\Sc,\Sb]$, the resulting image have a size of $[n \cdot \Sr,\Sc,\Sb]$.
The main advantage of this technique is that pixel interleave is unchanged (typically $row$, $column$, $channel$), guaranteeing the efficient copy and processing.

\subsubsection{Training}
\label{sss_AppTraining}

Since the high level \tf API is Python, we provide a light Python environment to allow developers to build and train their models from patches images produced using our sampling application described in Section \ref{sss_AppSampling}. As there is already a number of available \tf models and existing user-oriented opens-source applications to create \tf models, we chose not to focus it.

Our contribution for RS images deep net training is a user-oriented OTB application dedicated for training existing \tf models.
The application can import any existing \tf model. It can restore from file model variables before the training, or save them to disk after training. Thus this application can be used either to train a particular model from scratch, or perform fine tuning depending on the variables restored.
The training can be performed over one or multiple inputs given their corresponding \tf placeholders names and provide usual evaluation metrics. 

\subsubsection{Model serving}
\label{sss_AppModelServing}

We introduce a single OTB application dedicated to \tf model serving, that implements a pipeline with the filter described in Section \ref{ss_ModelFilter}.
It produces one output image resulting from \tf model computations.
As the entire pipeline implements the streaming mechanism described in Section \ref{ss_Pipelining}, it can process one or multiple images of arbitrary sizes.
The user can adjust the produced images blocks sizes, thanks to the internal OTB application architecture.
This application provides a generic support for the operational deployment of trained deep nets on RS images.

\subsubsection{Composite applications}
\label{sss_AppComposite}

Composite applications are OTB applications connected together.
Recent studies have suggested that deep nets features can be used as input features of algorithms like classification, leading state of the art results.
Regarding RS image classification, OTB already implement a number of  algorithms in its classification application, e.g. SVM, Random Forests, boost classifier, decision tree classifier, gradient boosted tree classifier, normal Bayes classifier.
In the sake of demonstrating the operational aspect of our approach, we implement a composite application that reuse our model serving application (described in \ref{sss_AppModelServing}) as input of two existing OTB applications: the \textit{TrainImagesClassifier} and \textit{ImagesClassifier} applications respectively dedicated to train a classifier and perform an image classification.

\section{Experiments}
\label{sec_Expe}

In this section, we analyze performances of our \tf model serving application.
We conduct a series of experiments on two representative state of the art deep nets for RS images, the Maggiori et al. fully convolutional model~\cite{MAGG} and the Ienco \mt data fusion model~\cite{IENC2}. 
Both perform land cover classification, but are different in terms of architecture as well as implementation. 
The model of Maggiori is a single-input CNN composed exclusively of convolutions, pooling and activation functions, allowing to process one image region with a single convolution.
The \mt model is an hybrid CNN-RNN model that inputs time series (TS) and one VHRS image. Unlike the Maggiori model, it consumes patches of the VHRS image and one pixel stack of time series, to perform a single estimation at one only location.
Table~\ref{tab_models} summarizes models parameters: Expression Field (EF), \disc{Receptive Fields (RF)}, mode and input reference image. Both models have an unitary scale factor.
\begin{table}[!t]
\renewcommand{\arraystretch}{1.3}
\caption{Models parameters}
\label{tab_models}
\centering
\begin{tabular}{cccc}
&
Maggiori FC17 
& 
Ienco \mt
\\
\hline
\disc{RF}
&
80x80 (\textit{VHRS})
& 
1x1 (\textit{TS}), 25x25 (\textit{VHRS})
\\
EF
&
16x16
& 
1x1
\\
Mode
&
Fully-conv
& 
Patch-based
\\
Reference
&
\textit{VHRS}
&
\textit{TS}
\\
\hline

\end{tabular}
\end{table}
The used images are presented in Table~\ref{tab_data}: one VHRS Spot-7 image acquired within the GEOSUD project\footnote{www.equipex-geosud.fr} and one S2 TS provided by the THEIA Land Data Center\footnote{http://www.theia-land.fr/en}. The TS consists in spectral bands resampled at 10m and additional radiometric indices.
The Maggiori net was trained to estimate two classes and \mt eight classes.
\begin{table}[!t]
\renewcommand{\arraystretch}{1.3}
\caption{Characteristics of the dataset}
\label{tab_data}
\centering
\begin{tabular}{|l|l|l|l|l}
\hline
Id & Product type & Size ($col \times row \times band$) & Pixel spacing\\
\hline
\textit{VHRS} & Spot-7 PXS 
&
\begin{tabular}{@{}c@{}}$24110 \times 33740 \times 4$ \end{tabular}
&
 1.5 meters\\
\textit{TS}  & S2 Time series 
&
\begin{tabular}{@{}c@{}}$3616 \times 5060 \times 592$  
\end{tabular}
&
 10 meters \\
\hline
\end{tabular}
\end{table}

\begin{figure}[!t]
\centering
\includegraphics[width=3.5in, height=1.86in]{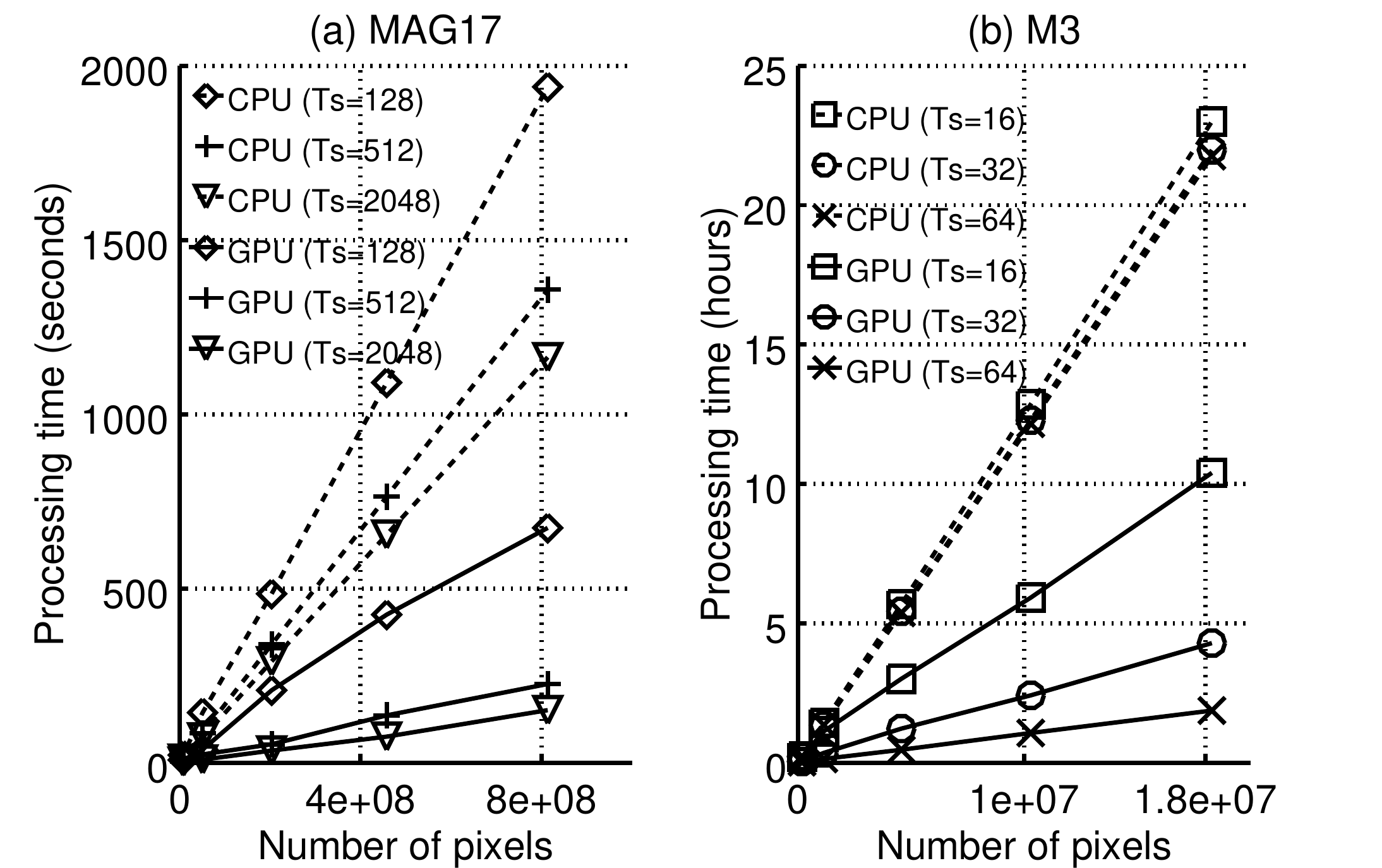}
\caption{Measured processing times for different images sizes for (a) the Maggiori model (MAG17) and (b) Ienco \mt model (\mt)}
\label{fig_pi_bench}
\end{figure}

\disc{All experiments are conducted on two computers.
The first is a basic workstation without GPU support (Intel Xeon CPU E5-2609 @ 2.50GHz and 16 GB of RAM). 
The second is a server with GPU support (Intel (R) Xeon (R) CPU E5-2667 v4@3.20Ghz with 256 GB of RAM and TITAN X GPU).
Libraries versions are OTB 6.7 and TF 1.7.}
We measure run times of our applications using the two models on both configurations, for different computed images sizes and various tile size. 
We presents separately the results for both models because their processing time differs largely due to the fact that \mt is more complex than the Maggiori model.
Figure~\ref{fig_pi_bench} shows that processing time is linear to the number of pixels of the produced image, regardless parameters.
Very large input datasets are entirely processed thank to the \textit{streaming} mechanism, even on the basic workstation with only 16Gb RAM.
\tf computations are benefiting strongly from the GPU support especially for image convolution, but CPU workstation processing time is reasonable with less than 24 hours for \mt (Fig.~\ref{fig_pi_bench}b) and less than 1 hour for the Maggiori model (Fig.~\ref{fig_pi_bench}a).

\section{Discussion}
\label{sec_disc}

\subsection{Operational deep learning framework for RS images}
Our first goal was to provide a generic DL framework for RS images processing. 
We propose a solution which takes roots in the OTB library. 
The original machine learning framework was enriched to embed \tf for training and using deep neural networks.
We introduce a new component enforcing the OTB workflow, that developers can implement in pipelines.
We present a set of \disc{new applications that users without programming knowledge can operate:} patches sampling, model training and serving, and also training and classification applications performing on features of any deep net exported as a \tf model.
Our approach is successfully applied to common RS images with two representative state of the art deep nets\disc{. Processing times measurements have shown a great \disc{scalability} and also that our applications can run deep nets without restriction on dataset size and regardless hardware configuration}.

\subsection{Limits and further research}
It should be noted that the presented method also has limits. 
The user has responsibility in providing crucial parameters to applications.
In particular, \disc{receptive field}, expression field and scale factor of deep nets.
In future development, those parameters could be extracted from the serialized \tf graph.
In the same way, the memory footprint is computed within the OTB pipeline, but does not takes in account the memory consumed internally by \tf during the execution of the computational graph: the user is thus responsible of fine tuning using the OTB application engine parameters.

\section{Conclusion}
This work was carried out with a view toward processing remote sensing (RS) images using deep learning (DL) techniques from the user perspective.
We propose a generic framework based on Orfeo Toolbox (OTB) and TensorFlow (\tf) libraries.
We have successfully applied existing state of the art deep nets on common RS images using our framework and shown a good computational efficiency\disc{, without restriction on images sizes and regardless hardware configuration.}
Our approach enables users without coding skills to use deep nets in their RS applications, and developers to create operational RS processing pipelines benefiting from the development framework of OTB library.
Our approach allows the combination of the existing OTB machine learning framework with deep nets.
The integration of DL processes in high performance computing architectures is enabled thank to the heterogeneous devices supported by the used libraries.
Further research could focus on improving the automatic retrieval of nets parameters and memory footprint. The source code \disc{and documentation} corresponding to the implementation presented in this paper is available at~\cite{GIT}, and the exposed framework will be proposed as official contribution in the forthcoming releases of OTB.

\section*{Acknowledgments}
The author thanks the OTB team, Dino Ienco and Emmanuel Maggiori.
This work was supported by public funds received through GEOSUD, a project (ANR-10-EQPX-20) of the \textit{Investissements d'Avenir} program managed by the French National Research Agency.
\discf{The author would also like to thank the reviewers for their valuable suggestions.}

\bibliographystyle{unsrt}
\bibliography{cresson2018}

\end{document}